\newif\ifisfinal\isfinaltrue \newif\ifisarxiv\isarxivtrue
\documentclass{article}

\usepackage[init=true]{calamrflow}
\ifisarxiv%
\usepackage[bibname=calamrflow,confparam={preprint}]{zvneurips}
\else
\usepackage[bibname=calamrflow,confparam={main}]{zvneurips}
\fi
\usepackage{zlsec}
\usepackage{zlfig}
\usepackage{zltable}
\usepackage{zllist}
\usepackage{zlmath}

\zaveninit%

\zeauthors%

\renewenvironment{zmwherelist}[1][, and\ ]
{where \begin{itemize*}[label={}, itemjoin={;\ }, itemjoin*={{#1}}, afterlabel={}, after={.}]}
{\end{itemize*}}

\newcommand{\zzsubsec}[2]{\subsect[#1]{#2}}
\newcommand{\zzpara}[2]{\para[#1]{#2}}

\newcommand{\tabsize}{\normalsize}
\usepackage{paper-table}

\begin{document}
\zavenbegindoc%
\begingroup
\renewcommand{\thefootnote}{*}
\footnotetext{These authors contributed equally.}
\endgroup

\begin{abstract}
  \Acp{llm} are optimized to produce distributionally plausible continuations
  rather than to explicitly verify whether generated propositions are entailed
  by source documents.
  This inductive bias enables generalization, but it does not encode whether
  responses are grounded with respect to a reference.
  These issues limit the use of \llms\ in domains where strict factual
  correctness is crucial, such as clinical decision support.
  Existing hallucination detection approaches improve factuality through
  retrieval augmentation, self-consistency, or claim verification, but generally
  do not learn directly over alignment topology.
  To leverage alignment topology as an inductive bias, we construct aligned
  bipartite graphs between reference information and \llm\ outputs and train a
  \gnnlongname{} (\acs{gnn}) to model alignment structure using message passing.
  The method achieves \sota\ results on four diverse hallucination and
  question-answering datasets, outperforming all compared methods, including
  foundational \llms\ such as GPT-4o.
\end{abstract}
\acresetall

\sect[intro]{Introduction}

Autoregressive \llms\ are trained to model likely next-token sequences rather
than to explicitly verify whether generated propositions are entailed by source
evidence, which can yield fluent but unsupported generations when explicit
grounding is required.  This behavior is consistent with prior work showing
that language models are trained with likelihood-based objectives and exhibit
inductive biases that shape generalization, but do not directly enforce
reference-based verification of generated
propositions~\cite{brownLanguageModelsAre2020a,holtzman2020curious,someyaInformationLocalityInductive2025,li2023inference}.
As a result, models prioritize fluency over grounded responses.

\pipelineOverview[t]

Methods for improving factuality and grounding are a popular area of research,
such as \zcpaperintro{rag}, which improves factuality by conditioning on
retrieved non-parametric memory.  Other methods, such as
SelfCheckGPT~\cite{manakulSelfCheckGPTZeroResourceBlackBox2023}, estimate
factuality from consistency across stochastic generations. However, these
approaches generally do not learn directly over the alignment
topology. Instead, they operate over retrieved passages, sampled text,
extracted claims, or external knowledge triples rather than an explicit
bipartite alignment graph.

Our method, which we call \method{}, contrasts these approaches by introducing
a structural inductive bias through graph-based alignment of reference and
response.  We construct linguistic graphs from prior context and align their
components using semantic similarity and a max-flow
algorithm~\cite{chenMaximumFlowMinimumCost2022,ford1962flows}. This formulation
encodes grounding as structured correspondences between reference and response,
allowing message passing to capture consistency and inconsistency in their
alignment.  The resulting alignment topology is then used to guide a \gnn,
which produces a probabilistic estimate of hallucination.

Our contributions are the following:
\begin{zlenumerateinline}
\item we introduce a graph-based grounding detector that learns over bipartite
  alignment topology
\item we demonstrate state-of-the-art performance across four diverse
  benchmarks
\item we provide analysis of alignment topology through graph rewiring and
  perturbation experiments
\item we show that perturbations to alignment structure measurably affect
  grounding performance across datasets
\ifisfinal%
\item we release \calamrsource[source code, configuration files, and
  reproduction instructions] to support replication of the reported experiments
\fi
\end{zlenumerateinline}

We posit that alignment topology provides a structural inductive bias for
grounding detection. The method achieves \sota\ performance on four
hallucination and question-answering benchmarks, including clinical and
biomedical settings.  It outperforms the reported automatic baselines across
these datasets and exceeds a reported single-annotator human macro-F1 value on
one benchmark while remaining below human accuracy.
\sect[related]{Related Work}

\para{Reference-grounded hallucination detection.}

Early evaluation of generated text often relied on surface-form overlap metrics
such as \rouge, which remain useful for measuring lexical similarity but do not
directly assess factual faithfulness. As language models have improved fluency
and coherence, evaluation has increasingly shifted from detecting malformed
output to identifying plausible but unsupported content. This distinction is
especially important in question-answering and clinical applications, where
fluent responses may still contradict or overstate the evidence available in
source documents.

Recent hallucination detection methods address this problem through
self-consistency~\cite{manakulSelfCheckGPTZeroResourceBlackBox2023},
reference-based claim
checking~\cite{huKnowledgeCentricHallucinationDetection2024}, and
knowledge-graph-based evaluation~\cite{sansford2024grapheval}.
RefChecker~\cite{huRefCheckerReferencebasedFinegrained2024} decomposes
responses into claim triplets and verifies them against references, while
GraphEval~\cite{sansford2024grapheval} represents information as
knowledge-graph triples to support more interpretable hallucination evaluation.
\rag\ addresses a related grounding problem by retrieving relevant passages and
conditioning generation on external evidence, thereby combining parametric
language-model knowledge with non-parametric sources.
Self-Reflection~\cite{jiMitigatingLLMHallucination2023} addresses hallucination
mitigation from a different angle by prompting an LLM to critique or revise its
own outputs, but it relies on model-internal judgments rather than explicit
verification against structured reference evidence.

\para{Question-answering and faithfulness evaluation.}

Question-answering provides a natural setting for evaluating factual
faithfulness because generated answers can be assessed against a question and
supporting reference evidence. Large-scale QA benchmarks such as
SimpleQuestions~\cite{bordesLargescaleSimpleQuestion2015} and
SQuAD~2.0~\cite{rajpurkarSQuAD1000002016} helped establish standard evaluation
settings for answer extraction and unanswerability. However, modern language
models can produce fluent free-form answers that are not necessarily entailed
by the provided context, motivating hallucination detection methods that
evaluate whether generated responses are supported by reference documents.
Biomedical question-answering sharpens this grounding problem because answers
must be supported by clinical or scientific evidence, as reflected in datasets
such as \pubmedqa{}.

\para{Grounded and traceable clinical generation.}

Clinical text generation has received increasing attention, building on both
general-purpose text-to-text modeling~\cite{raffelExploringLimitsTransfer2020}
and clinical factuality-oriented generation
methods~\cite{zhangOptimizingFactualCorrectness2020}. Unlike many open-domain
generation settings, clinical documentation must be both faithful to source
evidence and traceable to the underlying patient record or clinical
context~\cite{shingClinicalEncounterSummarization2021,zhangOptimizingFactualCorrectness2020}.
These requirements make tasks such as discharge-summary generation especially
challenging~\cite{adamsWhatSummaryLaying2021}. Recent work on generating
complete hospital discharge
summaries~\cite{landesCompleteHospitalDischarge2026} uses the same alignment
method adopted in our approach, \calamr, motivating our use of structured
alignment for clinical hallucination detection.

\para{Semantic graphs for grounded generation.}

\Ac{amr} represents sentence meaning as a graph rooted in natural language
semantics~\cite{\banarescuct}. Unlike knowledge graphs, which typically encode
curated relations among abstract entities or concepts, \amr\ graphs are derived
from text and represent predicate--argument structure, entities, attributes, and
relations expressed in a sentence. \Ac{amr} has been used as a structured
semantic representation for a range of language tasks, including
\ac{nlg}~\cite{manningHumanEvaluationAMRtoEnglish2020},
\acl{mt}~\cite{blloshmiXLAMREnablingCrossLingual2020}, and
\acl{qa}~\cite{limKnowWhatYou2020}. In our work, \amr\ graphs are generated
from reference context and aligned with generated output using \calamr, enabling
a graph-based comparison between source evidence and model response. This
positions semantic graph alignment as a structured alternative to LLM-internal
self-critique or purely lexical overlap for verifying whether a generated
response is grounded in reference evidence.

\sect[method]{Method}

We study reference-grounded hallucination detection as graph-level
classification over aligned semantic representations.  Given a reference text
and a candidate response, the model predicts whether the response is supported
by the reference.  Our approach first converts both texts into \ac{amr}-derived
semantic graphs, aligns semantically related nodes across the reference and
response, and classifies the resulting aligned graph with a
\zcpaperintromethod{gnn}.  The central hypothesis is that support and
hallucination are not only properties of individual lexical matches, but also
of the topology induced by cross-text semantic alignment.

\subsect[task-form]{Task Formulation}

Each example consists of a reference--response pair $(x,y)$ and a binary label
$z \in \{0,1\}$.  The reference text $x$ denotes the evidence, context, source
document, or other grounding material against which the candidate response $y$
is evaluated.  The response is either supported by the reference or contains
unsupported content.  For hallucination datasets, the positive class denotes
hallucinated or unsupported output.  For question-answering datasets, we cast
the task into the same reference-grounded classification form by treating the
provided context as $x$ and the generated answer as $y$, with the label
indicating whether the answer is supported by the context and correctly
addresses the question.  Thus, all datasets are reduced to a common
input--output interface: classify a reference--response pair from its aligned
semantic structure.

In the terminology of \calamr~\cite{\calamrct}, the reference text is treated
as the source and the candidate response is treated as the summary.  We use
this terminology only to reuse the existing graph construction and alignment
pipeline; the learning objective in this work is hallucination detection rather
than summary evaluation.

\subsect{Semantic Graph Construction}

We use \zcpaperintro{calamr} to construct semantic graph representations for
reference--response pairs\@.  \calamr{} was originally developed to align source
and summary texts.  Here, we adapt the same machinery to align reference texts
with generated responses.  For each input pair, the reference text is parsed
into a document-level \ac{amr} graph composed of sentence-level subgraphs.  The
candidate response is parsed in the same way.  Nodes correspond to semantic
concepts, attributes, and related graph elements, while edges encode the
predicate--argument and semantic relations induced by the \ac{amr} parse.

We adapt the \calamr{} alignment formulation to reference--response pairs.
Each node receives a semantic embedding derived from its sentence context,
aligned concept tokens, \zcpaperintro{propbank} role information, or attribute
value, depending on node type.  Alignment capacities are then computed from
cosine similarity between neighborhood-aware node representations.  The full
reference--response adaptation of the \calamr{} embedding, capacity, and
flow-network formulation is provided in \apxref{calamr:form}.

The two document graphs are then combined into a single aligned bipartite
graph.  Edges internal to each text preserve the original semantic structure,
while cross-text edges represent candidate alignments between reference and
response nodes.  These alignment edges are induced from embedding-based
semantic similarity over sentence, concept, and attribute nodes, with candidate
links retained when their capacity is positive after the similarity transform
defined in \secref{align-top}.  The resulting representation preserves both
within-text semantic structure and between-text correspondence, allowing the
classifier to reason over which parts of the response are grounded in the
reference.

\subsect[align-top]{Alignment Topology}

The aligned graph can be viewed as two semantic disconnected graph components
coupled with a set of cross-text alignment edges to form a bipartite graph.  We
refer to this cross-text bipartite graph as the alignment topology and use it
as the main representation for the downstream \gnn.  Intuitively, a faithful
response should induce coherent alignment between response concepts and
supporting reference concepts, while hallucinated content should appear as
weakly connected, sparsely aligned, or structurally inconsistent regions of the
graph.

\calamr{} assigns cross-text alignment scores using embeddings associated with
\ac{amr} sentence, concept, and attribute nodes, together with semantic role
information from \propbank.  Before computing alignment capacities, node
embeddings are expanded with local graph context so candidate matches depend on
nearby semantic structure rather than isolated node labels.  For reference node
$u \in V_x$ and response node $v \in V_y$, the alignment edge capacity is
computed from a bounded transformation of neighborhood-aware cosine similarity:
\begin{equation}
\begin{gathered}
\begin{aligned}
\sigmoid[x] &=
  \bigl(1 + \exp(0.5 - x)\bigr)^{-1} - 0.5,
\hspace{0.75em}
& c =\simfn[\simdampenmath_n]{\mathbf{h}_u}{\mathbf{h}_v}
\end{aligned}
\\
\zmfunc{C}_{u,v}(\mathbf{h}_u, \mathbf{h}_v)
\zmdef
\min\!\left(1, \max\!\left(0, c + \sigmoid[c]\right)\right). \nonumber
\end{gathered}
\end{equation}
where $C_{u,v}(\mathbf{h}_u,\mathbf{h}_v)$ is the \calamr{} alignment capacity
between nodes $u$ and $v$, $\mathbf{h}_u$ and $\mathbf{h}_v$ are
neighborhood-aware node embeddings, $\mu_n$ controls similarity dampening,
$\sigma$ is the translated sigmoid used by \calamr{} to emphasize stronger
alignments, $x$ is a hyperparameter that scales the sigmoid translation, and
$c$ is the dampened neighborhood-aware cosine similarity.  The full sentence,
concept, attribute, neighborhood, and capacity definitions are provided in
\apxref{calamr:form}.

The resulting aligned graph is
\[
G_{xy} = (V_x \cup V_y, E_x \cup E_y \cup E_{xy}),
\]
where $V_x,V_y$ are the reference and response nodes, $E_x,E_y$ are the
intra-text \ac{amr} edges, and $E_{xy}$ contains cross-text alignment edges
weighted by their \calamr{} capacities.

These scores are incorporated into a network-flow formulation that estimates
information flow across the aligned reference--response graph using the
max-flow algorithm~\cite{ford1962flows,chenMaximumFlowMinimumCost2022}.
Low-capacity or low-flow alignment edges downweight weakly supported
correspondences, whereas high-flow edges preserve semantically informative
links between the reference and response.  Local semantic context is captured
through neighborhood-weighted node representations, while global support
structure is captured by the flow values induced over the aligned graph.  The
result is a weighted alignment topology in which graph structure and edge
weights jointly encode evidence of support.

Unlike prior use of \calamr{} for source--summary evaluation, we use the
alignment topology as a supervised representation for hallucination detection.
The classifier is not given a free-text explanation or an \llm{} judgment.
Instead, it receives the structured graph induced by the reference, the
response, and their cross-text semantic alignments.  This design makes the
inductive bias explicit: the model is encouraged to learn patterns of grounded
and ungrounded correspondence rather than relying only on surface-form overlap
or parametric world knowledge.

\subsect{Graph Neural Classifier}

We classify each aligned graph with a \gnn.  Let $G=(V,E)$ denote the aligned
reference--response graph, where $V$ contains nodes from both texts and $E$
contains both intra-text semantic edges and inter-text alignment edges.  Each
node is represented by a 771-dimensional feature vector.  The first 768
dimensions are a \zcpaperintromethod{sbert} embedding of the node text.  The
remaining three dimensions encode \calamr{}-specific structural attributes: the
node type, whether the node belongs to the reference or response subgraph, and
whether the node corresponds to a \calamr{} concept.

The encoder is a multi-layer graph convolutional network that performs message
passing over the aligned graph~\cite{kipfSemiSupervisedClassificationGraph2017}.
Message passing allows each node representation to incorporate information
from its local semantic neighborhood and from cross-text alignment edges.  We
then apply attention-based graph pooling to obtain a graph-level
representation~\cite{liGatedGraphSequence2016}.  A feed-forward classifier maps
this pooled representation to a binary prediction indicating whether the
candidate response is supported by the reference.

This architecture is intentionally lightweight relative to the \llm{} baselines
used for comparison.  The model does not generate text and does not query an
external language model during inference.  Its prediction is based on the
semantic graph structure and the learned relationship between alignment
patterns and reference-grounded correctness.

\subsect{Training Objective and Implementation}

The graph encoder has three graph convolutional layers.  The input dimension is
771, followed by a hidden layer of size 256 and a final graph representation of
size 256.  The classifier head maps the pooled graph representation to a
128-dimensional hidden layer and then to a single logit.  We optimize binary
cross-entropy using \textsc{AdamW}~\cite{loshchilovDecoupledWeightDecay2018}
with learning rate $1 \times 10^{-3}$ and dropout $0.2$.  Models are trained
for up to 50 epochs, with convergence observed after nine epochs on
\medhallu{}.

All datasets are evaluated under the same binary classification formulation.
We report standard classification metrics, including precision, recall, and
F1, with the hallucinated or unsupported class treated as the positive class
unless otherwise stated.  This convention emphasizes the intended use case:
identifying outputs that should not be trusted without additional review.

\sect[experiments]{Experiments}

\subsect[exp-design]{Experimental Design}

\calamr\ was originally designed to operate over a pair of source and summary
texts by converting each text into an \amr-derived semantic graph and then
representing their relationship with cross-text alignment edges.  We adapt this
source--summary formulation to reference-grounded hallucination detection by
treating each input as a reference--response pair $(x,y)$: the reference $x$
contains the evidence, context, or knowledge source, and the response $y$ is the
candidate answer or generated text evaluated for support.

Under this formulation, question answering can be viewed as a structured
grounding problem.  The context and question define the reference, while the
answer plays the role of the response whose content must be supported by that
reference.  We hypothesize that the aligned graph representation gives the
\gnn\ access to structural evidence that distinguishes grounded answers from
unsupported or hallucinated ones.  This parallels how a human evaluator checks
whether a response accurately reflects the source material, but represents the
comparison as a learned graph-level classification problem.

Each aligned graph is used as one supervised learning instance.  The graph is
the model input, and the binary label indicates whether the response is
hallucinated or unsupported.  The model therefore learns to classify
reference--response alignment topology rather than relying on generated
explanations, surface-form overlap alone, or an \llm\ judge at inference time.

As defined in \secref{task-form}, all datasets are cast into the same
reference--response classification with binary label $z \in \{0,1\}$, where
$z=1$ denotes hallucinated or unsupported output and $z=0$ denotes supported
output.  The experiments therefore evaluate whether the aligned \calamr\ graph
topology contains sufficient structural evidence for the \gnn\ to distinguish
supported from unsupported responses across domains.

\subsect[datasets]{Datasets}

We evaluate on four datasets covering general-domain hallucination detection,
controlled hallucination detection, medical hallucination detection, and
biomedical question answering.

\introds{halueval} We use \halueval{} as a general-domain hallucination benchmark.
The provided context is treated as the reference text $x$, and the candidate
output is treated as the response $y$.  Dataset labels are mapped to the common
binary classification setting in which the positive class denotes hallucinated
or unsupported output.

\introds{hdm} We use \hdm{} to evaluate hallucination detection
under a controlled reference--response setting.  Each source passage is treated
as the reference, and the candidate answer or summary is treated as the
response.  This dataset tests whether alignment topology captures support
violations introduced by systematic perturbations of otherwise plausible
outputs.

\introds{medhallu} We use \medhallu{} to evaluate grounding detection in the
medical domain, where unsupported outputs may be fluent, domain-specific, and
difficult for non-experts to identify.  The medical question and supporting
context form the reference, while the generated answer forms the response.  The
model predicts whether the answer is supported by the supplied evidence.

\introds{pubmedqa} We use \pubmedqa{} as a biomedical question-answering
benchmark over PubMed abstracts.  Each instance contains a biomedical research
question, an abstract-derived context, and a gold yes/no/maybe answer.  We
treat the question and abstract context as the reference $x$ and the candidate
answer as the response $y$.  The example is labeled as supported when the
response agrees with the gold answer and is grounded in the abstract context.

\Tabref{datasetLabelDistTab} lists each dataset's statistics.

\datasetLabelDistTab{\tabsize}

\subsect[data-processing]{Data Processing}

All datasets are converted to the same graph construction format.  For each
example, \calamr\ processes the reference and response texts into separate
semantic graphs.  These graphs are then aligned by adding weighted cross-text
edges between semantically related nodes.  The resulting aligned graph preserves
both within-text semantic structure and between-text correspondence, allowing
the classifier to learn from the topology of support between the reference and
response.

This preprocessing step requires only the reference text, response text, and
binary label for each sample.  The generated graph is used directly as the
input to the \gnn.  No dataset-specific architecture changes are made, and no
free-text rationales or model-generated judgments are provided to the
classifier.

\subsect[gnn-training]{GNN Training}

We train one \method{} model per dataset using a 70\%/15\%/15\% train,
validation, and test split, and evaluate each setting over 5 random seeds. We
report mean test-set performance across seeds. Datasets are kept separate: a
model trained on one dataset is evaluated only on that dataset unless otherwise
specified. This prevents test-set leakage and allows us to assess whether the
same alignment-topology formulation is effective across domains. Baseline values
from prior work are reported as published and are not re-estimated across seeds.

The classifier is a graph convolutional network with attention-based graph
pooling.  It receives the aligned semantic graph as input and produces a binary
prediction indicating whether the response is hallucinated or unsupported.  We
train with binary cross-entropy and report precision, recall, and F1 with the
hallucinated or unsupported class treated as the positive class.

The processed datasets vary in both size and class balance.  \medhallu{}
and \halueval{} are approximately balanced between supported and
hallucinated examples, whereas \hdm{} is skewed toward hallucinated
examples and \pubmedqa{} is skewed toward supported examples.  Because
of these differences, we report precision, recall, and F1 for the hallucinated
or unsupported class rather than relying on accuracy alone.

\subsect[training-details]{Training Details}

We use the same graph neural architecture and optimization procedure across all
datasets.  The graph encoder has three message-passing layers with hidden
dimension 256 and dropout 0.2.  Graph-level representations are computed with
attention pooling using a $\tanh$ attention activation.  The classifier maps
the pooled graph representation to a single logit, which is optimized with
binary cross-entropy.  Following the task formulation in \secref{task-form},
the target label is $z=1$ for hallucinated or unsupported responses and $z=0$
for supported responses, including correct answers in question-answering
datasets.

We optimize with \textsc{AdamW}~\cite{loshchilovDecoupledWeightDecay2018} using
learning rate $10^{-3}$, weight decay $10^{-5}$, and mini-batches of size 32.
Training runs for at most 50 epochs.  The learning rate is reduced by a factor
of 0.5 when validation loss fails to improve for three consecutive epochs.
Early stopping is also based on validation loss, with patience 5 and minimum
improvement $\delta=0.001$.  Gradients are clipped to global norm 1.0 during
each training epoch.

The number of epochs completed before early stopping varies by dataset:
\medhallu{} converges after 15 epochs, \halueval{} after 12 epochs, \hdm{}
after 21 epochs, and \pubmedqa{} after 9 epochs.

\datasetResultsTab{\tabsize}

\sect[results]{Results}

\para{Predictive performance.}

\Tabref{datasetResultsTab} reports test-set performance across four datasets,
ordered from broad hallucination detection to increasingly domain-specific
reference--response evaluation. \halueval{} provides a large general hallucination
benchmark, \medhallu{} provides a large clinical hallucination benchmark, \hdm{}
evaluates medical hallucination detection with reported system baselines, and
\pubmedqa{} tests biomedical question answering.

\method{} achieves the strongest reported performance on the two largest
hallucination benchmarks, reaching 94.8\% Hallucination F1 on \halueval{} and
89.2\% on \medhallu{}. On \hdm{}, \method{} improves Hallucination F1 over \hdm{}-2
by 11.4 points while maintaining comparable accuracy. On \pubmedqa{}, \method{}
exceeds the single-annotator human macro-F1 baseline but remains below human
accuracy, suggesting more balanced class-level performance while leaving room
for improvement in exact decision accuracy.

\para{Graph-structure intervention analysis.}

To evaluate whether \method{} uses reference--response topology rather than
only node-level text features, we intervene on the alignment graph after
construction. We consider two intervention types: edge removal, which deletes a
controlled fraction of alignment connectivity, and edge rewiring, which
reassigns alignment edges while preserving the edge budget. We report both
perturbation and ablation settings to distinguish test-time corruption from
controlled graph reconstruction.

\ablPertMeanPlot{}

\Figref{ablPertMeanPlot} shows that perturbation produces the clearest topology
sensitivity. Hallucination F1 declines as increasing fractions of edges are
removed, and edge rewiring produces a smaller but consistent decrease,
indicating that the model is sensitive not only to the number of alignment
edges but also to their placement. The ablation setting is less monotonic:
performance varies with the amount of structural modification rather than
degrading uniformly. This suggests that some reconstructed graphs can partially
compensate for removed or reassigned edges, while still indicating that the
learned decision function is affected by alignment topology.

Full precision, recall, macro-F1, and accuracy curves for these interventions
are provided in \apxref{res:structure-interventions}.  Full predictive metrics
are reported in \apxref{full-predictive-metrics}.

\para{Semantic-similarity confound.}

We also test whether performance can be explained by raw semantic similarity
rather than graph-structured message passing. Mean aligned cosine distances are
similar for truth and hallucination examples at the \calamr{} input layer, but
separation changes across learned \gnn{} layers. Thus, the model does not simply
threshold input-level embedding similarity; it reshapes alignment-distance
distributions through message passing over the reference--response graph. Full
cosine-distance plots are provided in \apxref{cosine-distance}.

\sect[discussion]{Discussion}

The results suggest that explicit reference--response graph structure provides
a useful inductive bias for hallucination detection. \method{} performs strongly
on the two largest hallucination benchmarks and remains competitive on smaller
biomedical evaluation settings. The perturbation experiments show that corrupting
edge structure reduces performance, supporting the claim that the detector uses
graph topology rather than only node-level semantic features.

This structure is most useful where detection must be auditable. Because
\method{} exposes an aligned reference--response topology, failures can be
analyzed in terms of missing, weak, or misplaced correspondences rather than only
as black-box classifier errors. This property is especially relevant for clinical
and biomedical applications, where unsupported generated content must be traced
back to supporting evidence.

The method also has limitations. It depends on semantic graph construction,
alignment, and flow-based edge weighting; errors in these upstream steps can
propagate to the classifier. The ablation curves are not uniformly monotonic,
suggesting that some reconstructed graphs compensate for removed or reassigned
edges. Deployment in clinical decision-support settings would require
prospective validation, calibration, and human review.

\sect[conclusion]{Conclusion}

We introduced \method{}, a reference-grounded hallucination detector that
represents each reference--response pair as an aligned, flow-weighted graph and
learns over this topology with a graph neural network. Across four datasets,
\method{} achieves strong hallucination-positive F1, including the strongest
reported performance on the two largest hallucination benchmarks evaluated.
Graph-structure interventions and semantic-similarity analysis indicate that
the model uses alignment topology learned through message passing rather than
raw embedding similarity alone.

\clearpage
\zavenenddoc%

\clearpage
\onecolumn
\zavenappendix%
\apx[res:structure-interventions]{Perturbation and Ablation Details}

\ablPertFullPlot[h!]{}

\clearpage
\apx[cosine-distance]{Cosine Distance Analysis}

\cosineDistPlotFig[h!]

\clearpage
\apx[tsne]{Embedding Visualization}

\embeddingPlotFig[h!]{\textwidth}

\clearpage
\apx[full-predictive-metrics]{Full Predictive Metrics}

\datasetResultsApxTab[h!]{\tabsize}

We evaluate on existing benchmark datasets and do not introduce a new dataset.
For each dataset, we use the dataset-provided label semantics and map the
binary target to \(z=1\) for hallucinated or unsupported output and \(z=0\) for
supported output. Hallucination F1, precision, and recall are therefore
positive-class binary metrics. Macro-F1 and accuracy are reported separately to
show class-balanced and aggregate performance. Baseline values are taken from
the corresponding dataset or model reports where indicated; all \method{}
results are computed on the test sets reported in \Tabref{datasetResultsTab}.

\apx{Compute Resources}

Experiments were run on GPU compute nodes with eight GeForce RTX 2080 Ti
Rev\@. A, with 12 GB VRAM, 40 CPUs, and 170 GB RAM\@.  Training one
dataset-specific model required approximately 1 day per seed, and each dataset
was evaluated over 5 random seeds. Graph construction and intervention analyses
are included in this estimate.

\apx[calamr:form]{Reference--Response \calamr{} Formulation}

Given a reference--response pair $(x,y)$, \calamr{} constructs an \amr{} graph
for the reference text $x$ and an \amr{} graph for the candidate response $y$.
We denote these graphs by $\zmgraph{G}_x=(\zmset{V}_x,\zmset{E}_x)$ and
$\zmgraph{G}_y=(\zmset{V}_y,\zmset{E}_y)$.  Alignment edges
$\zmset{E}_a \subseteq \zmset{V}_x \times \zmset{V}_y$ connect semantically
related nodes across the two components.  The formulas below define the node
embeddings and alignment capacities used to construct this weighted aligned
reference--response graph.

\zzsubsec{calamr:emb}{Node Embeddings}

\zzpara{calamr:emb:doc}{Document nodes.}

Document nodes aggregate over their child sentence and subgraph nodes:
\begin{align}
\emb{d} \zmdef \frac{1}{|\childnodemath[\zmst{n}]|}
  \sum_{\zmst{u} \in \childnodemath[\zmst{n}]} \emb[u]{} .\label{eq:app-calamr-doc}
\end{align}
\begin{zmwherelist}
\item $\zmst{n}$ is a document node from either the reference graph or response
  graph
\item $\childnodemath[\zmst{n}]$ is the set of child nodes of $\zmst{n}$
\item $\emb[u]{}$ is the embedding of child node $\zmst{u}$
\end{zmwherelist}

\zzpara{calamr:emb:sent}{Sentence nodes.}

Sentence nodes use the \sbert{} sentinel embedding of the sentence text:
\begin{align}
\emb{s} \zmdef \embfn{s_n} \in \zmrealdim[d] .\label{eq:app-calamr-sent}
\end{align}
\begin{zmwherelist}
\item $\zmst{n}$ is a sentence node in either component graph
\item $\embfn{s_n}$ is the \sbert{} sentinel embedding for sentence
  $\zmst{s_n}$
\item $d$ is the \sbert{} embedding dimension
\end{zmwherelist}

\zzpara{calamr:emb:tok}{Token-aligned concept component.}

For concept nodes, token-aligned text contributes an embedding from the final
\sbert{} output layer.  When no token alignment is available, a constant vector
is used so the node remains defined:
\begin{align}
\emb{t} & \zmdef
\begin{cases}
  \mathlarger{\sum_{\zmst{w}}^{\zmst{s_n}}\sum_{i}^{|\zmst{w}|} \embfn{s_n}_i} &
    \parbox{1.7cm}{if $\zmst{n}$ has \\ alignments} \\
  \zmvec{1_\mathit{d}} & \text{otherwise}
\end{cases} .\label{eq:app-calamr-concepttoks}
\end{align}
\begin{zmwherelist}
\item $\zmst{n}$ is a concept node
\item $\zmst{w}$ ranges over text tokens aligned to $\zmst{n}$ in sentence
  $\zmst{s_n}$
\item $\embfn{s_n}_i$ is the \sbert{} vector for the $i$th word-piece token
\item $d$ is the \sbert{} embedding dimension
\end{zmwherelist}

\zzpara{calamr:emb:role}{PropBank roles.}

A \pbroleset{} supplies semantic role information for verb concept nodes.  The
role embedding combines the role name, role semantic tag, and child-node
embedding:
\begin{align}
\tmpvec{} = \, & [ \embfn{s_r} + \embfn{s_f} ] \cdot \rlrolewmath \nonumber \\
\emb[e]{r} \zmdef \, & \embnobm[\childnodemath]{} \cdot \rlchildwmath + \tmpvec{} .\label{eq:app-calamr-conceptrole}
\end{align}
\begin{zmwherelist}
\item $\zmst{e}$ is the \pbrole{} edge
\item $\embfn{s_r}$ and $\embfn{s_f}$ are the \pbrole{} name and semantic-tag
  embeddings
\item $\embnobm[\childnodemath]{}$ is the embedding of the \pbrole{} child node
\item \rlchildwmath{} and \rlrolewmath{} are role and role-child
  \hyperparams{}
\end{zmwherelist}

\zzpara{calamr:emb:concept}{Concept nodes.}

The full concept-node embedding aggregates token alignments, \pbroleset{}
information, and outgoing role-edge embeddings:
\begin{align}
\tmpvec{rs} & =
  \1[v]{\zmst{n}} \, \embfn{s_{rs}} + (1 - \1[v]{\zmst{n}}) \, \embfn{s_n}
  \nonumber \\
\tmpvec{r} & =
  \1[v]{\zmst{n}}
  \left[
    \sum_{\zmst{e} \in \zmnetneigh[n]}\emb[e]{r} \, \cnptrlwmath
  \right] \nonumber \\
\emb{c} & \zmdef
  \emb{t} \, \cnpttokwmath + \tmpvec{rs} \, \cnptrswmath + \tmpvec{r} .\label{eq:app-calamr-concept}
\end{align}
\begin{zmwherelist}
\item $\emb{t}$ is the token-alignment embedding in
  \zmeqref{app-calamr-concepttoks}
\item $\embfn{s_{rs}}$ is the \sbert{} \pbroleset{} embedding
\item $\embfn{s_n}$ is the non-verb node text embedding
\item $\zmnetneigh$ are the outgoing role edges of node $\zmst{n}$ that
  connect to its children
\item \cnpttokwmath{}, \cnptrswmath{}, and \cnptrlwmath{} are \hyperparams{}
\item $\tmpvec{r}$ is the \pbrole{} contribution in
  \zmeqref{app-calamr-conceptrole}
\end{zmwherelist}

\zzpara{calamr:emb:attrib}{Attribute nodes.}

Attribute nodes use token alignments when available and otherwise fall back to
the attribute surface text:
\begin{align}
\tmpvec{t} = & \min(1, |\zmset{T}|) \, \emb{t} \nonumber \\
\emb{a} = \tmpvec{t} + & (1 - \min(1, |\zmset{T}|)) \, \embfn{s_n} .\label{eq:app-calamr-attribute}
\end{align}
\begin{zmwherelist}
\item $\zmset{T}$ is the set of token alignments for the attribute node
\item $\emb{t}$ is the token-alignment embedding in
  \zmeqref{app-calamr-concepttoks}
\item $\embfn{s_n}$ is the attribute surface-text embedding
\end{zmwherelist}

\zzpara{calamr:emb:netneigh}{Network neighborhood.}

Before cross-component matching, node embeddings are expanded with local graph
context.  Let $\kthorderfn{n}$ be the $k$th-order nodes at exactly $k$ hops from
node $\zmst{n}$.  The network-neighborhood embedding is:
\begin{align}
\emb{n} = \sum_i^k \sum_{\zmst{u} \in \kthorderfn[i]{n}}
  \emb[u]{} \cdot \kthorderwmath_i
  \label{eq:app-calamr-netneigh} \\
\forall x \in \kthorderwmath : x > 0 . \nonumber
\end{align}
\begin{zmwherelist}
\item $k$ is the maximum neighborhood order
\item $\kthorderfn[i]{n}$ is the set of nodes exactly $i$ hops from $\zmst{n}$
\item $\emb[u]{}$ is the embedding of neighbor node $\zmst{u}$
\item $\kthorderwmath_i$ is the $i$th-order weight \hyperparam{}
\end{zmwherelist}

\zzsubsec{calamr:cap}{Alignment Capacities}

Cross-component candidate pairs are scored by cosine similarity and then
nonlinearly dampened:
\begin{align}
\cossimfn & =
  \Biggl[
    \frac{\emb[n_1]{} \cdot \emb[n_2]{}}
         {||\emb[n_1]{}|| \, ||\emb[n_2]{}||}
   \Biggr] \nonumber \\
\simfn{n_1}{n_2} & =
  \cossimfn^\simdampenmath, \quad \simdampenmath > 0 .
  \label{eq:app-calamr-simfn}
\end{align}
\begin{zmwherelist}
\item $\emb[n_1]{}$ and $\emb[n_2]{}$ are embeddings of a candidate
  reference--response node pair
\item \simdampenmath{} is a node-type-specific \hyperparam{} that adjusts the
  similarity score
\end{zmwherelist}

The dampened similarity is transformed into a bounded alignment capacity:
\begin{align}
\sigmoid & = \bigl( 1 + \exp (0.5 - x) \bigr)^{-1} - 0.5
  \nonumber \\
c & = \simfn[\simdampenmath_n]{n_1}{n_2} \nonumber \\
\capfn{n} & \zmdef \min(1, \max(0, c + \sigmoid[c])) .\label{eq:app-calamr-capdef}
\end{align}

\zzpara{calamr:cap:doc}{Document-node capacities.}

Document-node alignment capacities are computed from document-node embeddings
using \zmeqref{app-calamr-capdef}:
\begin{align}
\capfn{d} \zmdef \capfn{n}(d_x,d_y) .\label{eq:app-calamr-doccap}
\end{align}
\begin{zmwherelist}
\item $d_x$ and $d_y$ are the reference and response document nodes
\end{zmwherelist}

\zzpara{calamr:cap:sent}{Sentence-node capacities.}

Sentence capacities are computed using \zmeqref{app-calamr-capdef}.  A
sentence-skew term linearly dampens concept and attribute capacities by the
compatibility of their parent sentence pair:
\begin{align}
\sentskewmath \zmdef
  \capfn{s} \, \sentdampmath + (1 - \sentdampmath),
  \quad \sentdampmath \in [0,1] .\label{eq:app-calamr-sentskew}
\end{align}

\zzpara{calamr:cap:concept}{Concept-node capacities.}

Concept alignment capacities are scaled by the sentence-skew term when the
candidate concept pair descends from an aligned sentence pair:
\begin{align}
\{ (\zmst{n_1}, \zmst{n_2}) \in \zmset{E}_a,
   (\zmst{s_1}, \zmst{s_2}) \in \zmset{E}_a \mid \nonumber \\
 \zmst{n_1} \in \descendsmath[\zmst{s_1}] \land
 \zmst{n_2} \in \descendsmath[\zmst{s_2}] \}, \nonumber \\
\capfn{c} \zmdef \capfn{} \cdot \sentskewmath .\label{eq:app-calamr-conceptcap}
\end{align}
\begin{zmwherelist}
\item $\zmset{E}_a$ is the set of alignment edges
\item $(\zmst{n_1},\zmst{n_2})$ is a candidate aligned concept-node pair
\item $(\zmst{s_1},\zmst{s_2})$ is the corresponding aligned sentence-node pair
\end{zmwherelist}

When aligned concept nodes share the same AMR variable name, as in reentrancies
or co-reference links across the reference and response components, the capacity
is set to its maximum value.

\zzpara{calamr:cap:attrib}{Attribute-node capacities.}

Attribute nodes have no AMR variable, so their capacities use the direct bounded
capacity and the sentence-skew term:
\begin{align}
\capfn{a} \zmdef \capfn{} \cdot \sentskewmath .\label{eq:app-calamr-attributecap}
\end{align}

\zzsubsec{calamr:flow}{Aligned Flow Network}

The weighted reference--response graph can be represented as a flow network by
adding a source node $\zmst{s}$ and sink node $\zmst{t}$ to the two AMR
components:
\begin{align}
\zmgraph{F}_{xy} \zmdef
  (\zmset{V}_x \cup \zmset{V}_y \cup \{\zmst{s},\zmst{t}\},
   \zmset{E}_x \cup \zmset{E}_y \cup \zmset{E}_a \cup
   \zmset{E}_{\zmst{s}} \cup \zmset{E}_{\zmst{t}},
   \capfn{}) .\label{eq:app-calamr-flowgraph}
\end{align}
\begin{zmwherelist}
\item $\zmset{E}_x$ and $\zmset{E}_y$ are intra-text semantic edges in the
  reference and response graphs
\item $\zmset{E}_a$ is the set of cross-text alignment edges
\item $\zmset{E}_{\zmst{s}}$ and $\zmset{E}_{\zmst{t}}$ connect the AMR
  components to the source and sink nodes
\item $\capfn{}$ assigns edge capacities using the formulas above
\end{zmwherelist}

A feasible flow $f$ satisfies edge-capacity and flow-conservation constraints:
\begin{align}
0 \le f(\zmst{u},\zmst{v}) \le \capfn{\zmst{u},\zmst{v}}
  \quad & \forall (\zmst{u},\zmst{v}) \in \zmset{E} \nonumber \\
\sum_{\zmst{u}:(\zmst{u},\zmst{v})\in\zmset{E}} f(\zmst{u},\zmst{v}) =
\sum_{\zmst{u}:(\zmst{v},\zmst{u})\in\zmset{E}} f(\zmst{v},\zmst{u})
  \quad & \forall \zmst{v} \notin \{\zmst{s},\zmst{t}\} .\label{eq:app-calamr-flowconstraints}
\end{align}

The maximum-flow objective is:
\begin{align}
\max_f |f| \zmdef
\max_f \sum_{\zmst{v}:(\zmst{s},\zmst{v})\in\zmset{E}}
  f(\zmst{s},\zmst{v}) .\label{eq:app-calamr-maxflow}
\end{align}

In this work, the supervised classifier consumes the aligned topology induced by
these embeddings and capacities.  The flow-network formulation explains how the
alignment procedure suppresses low-capacity correspondences and preserves
high-capacity reference--response structure.

\ifisarxiv%
\fi

\end{document}